# Prognosis Of Lithium-Ion Battery Health with Hybrid EKF-CNN+LSTM Model Using Differential Capacity

Md Azizul Hoque[1], Babul Salam[2*], Member, IEEE, Mohd Khair Hassan[1], Member, IEEE, Abdulkabir Aliyu[3], Abedalmuhdi Almomany[2], Member, IEEE, and Muhammed Sutcu[2], Member, IEEE

[1]Department of Electrical and Electronic Engineering, Universiti Putra Malaysia,43400, UPM, Serdang, Selangor, Malaysia
[2] Electrical Engineering Department, Gulf University of Science & Technology, Hawally,32093, Kuwait
[3] Faculty of Mechanical Technology and Engineering, Universiti Teknikal Malaysia Melaka, Durian Tunggal, Melaka, Malaysia.

Corresponding author: Second A. Author (babul.s@gust.edu.kw).

"This This project has been partially supported by Gulf University for Science and Technology and the GUST Engineering and Applied Innovation Research (GEAR) Center under project code: ISG – Case 61.''

**ABSTRACT** Battery degradation is a major challenge in electric vehicles (EV) and energy storage systems (ESS). However, most degradation investigations focus mainly on estimating the state of charge (SOC), which fails to accurately interpret the cells' internal degradation mechanisms. Differential capacity analysis (DCA) focuses on the rate of change of cell voltage about the change in cell capacity, under various charge/discharge rates. This paper developed a battery cell degradation testing model that used two types of lithium-ions (Li-ion) battery cells, namely lithium nickel cobalt aluminium oxides (LiNiCoAlO2) and lithium iron phosphate (LiFePO4), to evaluate internal degradation during loading conditions. The proposed battery degradation model contains distinct charge rates (DCR) of 0.2C, 0.5C, 1C, and 1.5C, as well as discharge rates (DDR) of 0.5C, 0.9C, 1.3C, and 1.6C to analyze the internal health and performance of battery cells during slow, moderate, and fast loading conditions. Besides, this research proposed a model that incorporates the Extended Kalman Filter (EKF), Convolutional Neural Network (CNN), and Long Short-Term Memory (LSTM) networks to validate experimental data. The proposed model yields excellent modelling results based on mean squared error (MSE), and root mean squared error (RMSE), with errors of less than 0.001% at DCR and DDR. The peak identification technique (PIM) has been utilized to investigate battery health based on the number of peaks, peak position, peak height, peak area, and peak width. At last, the PIM method has discovered that the cell aged gradually under normal loading rates but deteriorated rapidly under fast loading conditions. Overall, LiFePO4 batteries perform more robustly and consistently than (LiNiCoAlO2) cells under varying loading conditions.

## I. INTRODUCTION

The demand for sustainable energy solutions, particularly efficient and reliable energy storage systems, has expanded enormously [1,2]. With the rapid adoption of electric vehicles (EVs) and the expansion of renewable energy sources, energy storage systems (ESS) have become critical in modern power grids. Among energy storage technologies, lithium-ion (Li-ion) batteries have emerged as the preferred choice due to their high energy density, extended cycle life, and low self-discharge rate[3,4].However, despite their diverse applications, the complexity of understanding and predicting the internal health of Li-ion batteries under various operational conditions presents considerable challenges [5], particularly in applications requiring high performance and safety. Recent research is mostly focused on state of charge (SOC) estimation and capacity readings, which, while significant, frequently provide limited insights into the batteries' real internal health [6]. This limitation is particularly evident in solid-state batteries, where conventional state of charge and capacity measurement techniques can be inaccurate due to the distinctive characteristics of solid electrolytes [7,8]. The absence of in-depth research into the internal health deterioration mechanisms of Li-ion batteries under operating stresses creates a crucial gap in knowledge needed to improve their durability and reliability. Understanding these degradation processes is critical not just for improving battery management systems, but also for maintaining batteries' long-term capability in demanding applications like EVs and grid storage.

Battery deterioration is a complicated and diverse process influenced by a variety of parameters such as cell chemical composition, operational stress, temperature fluctuations, and the frequency and intensity of charging and discharging cycles [9,10,11]. The decline in a battery's ability to store and release energy over age leads to diminished performance and eventual failure [12,13]. This research examines the internal health of two varieties of Li-ion battery cells: Li(NiCoAl)O$_2$ and LiFePO$_4$. These cells were selected for their varying chemical compositions and performance attributes, offering a thorough understanding of the multiple degradation routes that occur under different operational conditions.

Diagnostic techniques such as Incremental Capacity Analysis (ICA) [14] and Differential Capacity Analysis (DCA) [15] enable the monitoring and assessment of battery degradation by detecting differences in capacity-voltage curves that reflect aging processes and performance decline. Identifying and examining these capacity peaks is essential for comprehending the fundamental deterioration mechanisms and predicting the battery's remaining useful life (RUL) [16]. The goal of this research is to discover how various degradation mechanisms influence battery performance decay and overall lifespan. This research approach addresses the technical aspects of battery degradation and the broader implications for optimizing charging cycles, reducing degradation rates, and boosting battery performance. DCA mathematically calculated as:

$$\left[\frac{dQ}{dV} \approx \frac{Q_{(k+1)} - Q_k}{V_{(k+1)} - V_k}\right] \tag{1}$$

Where $dQ$ represents the incremental change in charge capacity, and $dV$ denotes the corresponding voltage change. This method provides insights into phase transitions and internal electrochemical behavior during charge-discharge cycles. Meanwhile, $Q_k$ and $Q_{(k+1)}$ are the charge capacities at each voltage step $V_k$ and $V_{(k+1)}$. The $\frac{dQ}{dV}$ curve is used to identify peak shifts, capacity fade, and resistance variations, enabling accurate battery health prognosis and degradation assessment.

Accurate battery health prediction requires the ability to limit the inherent noise in battery data and understand the intricate interrelationships between various parameters such as current, voltage, capacity, and temperature [17,18]. Several studies utilize deep learning and machine learning techniques for predicting battery health [19,20]. Despite its outstanding precision, it has not succeeded in capturing the internal behavior of the battery during charge and discharge cycles. This research utilizes a hybrid modeling strategy that combines an extended Kalman filter with machine learning techniques, namely a Convolutional Neural Network (CNN) and a Long Short-Term Memory (LSTM) network to ingrate advanced learning techniques for accurate health prognosis. The Kalman filter lowers noise and increases data reliability [21]. The CNN and LSTM algorithms are used to extract spatial and temporal dependencies from battery data [22,23]. This hybrid technique increases battery health prediction accuracy by accounting for both long-term trends and short-term changes in battery performance data.

Furthermore, peak detection in differential capacity analysis is critical for diagnosing and predicting battery health [24]. Investigating variations in differential capacity facilitates the identification of nuanced degradation patterns sometimes ignored by conventional methods. This method is particularly advantageous for real-time health monitoring in dynamic applications, where batteries experience fluctuating loads and environmental conditions.

Key contributions of the paper:
1. Proposed a hybrid model that integrates the Extended Kalman Filter (EKF), Convolutional Neural Networks (CNN), and Long Short-Term Memory (LSTM) for predicting lithium-ion battery health with an accuracy of 99.99% under load conditions.
2. Evaluated the model's efficacy across various C-rates, revealing enhanced predictive precision at elevated C-rates owing to expedited convergence and reduced capacity.
3. Differential capacity analysis discovered phase transition shifts, peak broadening, and overpotential increases, providing a deeper understanding of lithium plating and structural fatigue in high C-rate scenarios.
4. Identified that LiFePO$_4$ demonstrated greater structural stability and longer cycle life, while LiNiCoAlO$_2$ showed increased polarization effects and faster degradation at higher C-rates.

## II. BATTERY MODEL

### A. EXTENDED KALMAN FILTER

The Extended Kalman Filter (EKF) is a recursive technique employed to estimate the state of a nonlinear system [25]. It enhances the conventional Kalman Filter by linearizing nonlinear functions about the current state estimate. It's implemented in the battery model to handle nonlinearities and data noise [26]. EKF comprises two primary phases: prediction and update. The process commences with the initialization of the state estimate and covariance. where the nonlinear functions are approximated by their first-order Taylor expansions which are shown in Figure 1. The main steps are stated below:

$$\hat{x}_{k|k-1} = f(\hat{x}_k - 1|k - 1, u_k - 1) \quad (2)$$
$$F_k = \frac{\Delta f}{\Delta x}|(\hat{x}_k - 1|k - 1, u_k - 1) \quad (3)$$
$$P_{k|k-1} = F_k P_{(k-1|k-1} F_k^T) + Q_k \quad (4)$$
$$H_k = \frac{\Delta h}{\Delta x}|(\hat{x}_k - 1|k - 1, u_k - 1) \quad (5)$$
$$K_k = P_{(k|k-1} H_k^T)(H_K P_{k|k-1} H_k^T + R_k)^{-1} \quad (6)$$
$$y_k = z_k h(\hat{x}_{k|k-1}) \quad (7)$$
$$\hat{x}_{k|k} = \hat{x}_{k|k-1} + K_k y_K \quad (8)$$
$$P_{k|k} = (I - K_k H_k) P_{k|k-1} \quad (9)$$

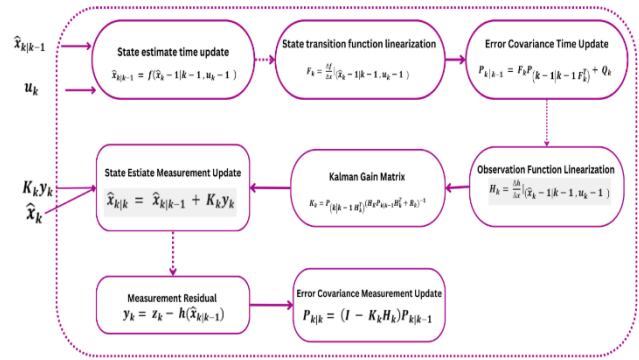

**FIGURE 1.** Block diagram representation of the extended Kalman filter

Initially, the state estimate $\hat{x}_0$ is set based on initial measurements, and the state covariance matrix $P_0$ is initialized to represent the initial uncertainty in the state estimate. In the prediction step, the state estimate is updated to the next time step using the nonlinear state transition function $f$, as expressed in equation 1 where $u_k - 1$ represents the control input from the previous time step. In the battery model, the control input is set as voltage. The state transition function is then linearized around the current estimate to obtain the Jacobian matrix in Equation 2. Subsequently, the state covariance matrix is updated to account for the process noise $Q_k$ which is stated in equation 3. The observation function linearizes the predicted state estimate, yielding the Jacobian matrix in equation 4. The Kalman gain $K_k$, is then computed to balance the uncertainty between the predicted state and the new measurement, given by equation 5, where $R_k$ represents the measurement noise covariance.

The measurement residual, the difference between the actual measurement $z_k$ and the predicted measurement $y_k$. This residual is used to update the state estimate, incorporating the measurement residual bias. Finally, the state covariance matrix is updated to reflect the new estimate $P_{k|k}$. The EKF iterates through this prediction and updates steps with each new measurement and control input, continuously refining the state estimate and covariance. This methodology ensures efficient and accurate state estimation for nonlinear systems by linearizing the nonlinear functions and maintaining a balance between prediction and measurement in this battery model.

### B. LONG SHORT-TERM MEMORY

Long Short-Term Memory (LSTM) networks, a specialized subset of recurrent neural networks (RNNs), are highly effective in modeling sequential data by capturing both spatial and temporal dependencies [27]. In battery datasets, LSTMs play a crucial role in maintaining spatial and temporal relationships across charge-discharge cycles [28]. This capability is essential for enhancing prediction accuracy and identifying degradation patterns throughout the battery's cycle life. The LSTM network relies on memory cells as its fundamental computational units, with each cell incorporating input, forget, and output gates. These gates regulate information flow, allowing the model to selectively retain critical features while mitigating the effects of short-term memory loss commonly observed in standard recurrent neural networks in battery modeling. Figure 2 illustrates the structural composition of the LSTM model.

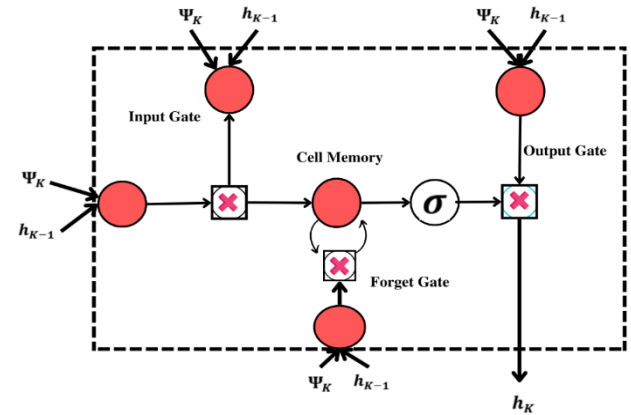

**FIGURE 2.** Structure of the LSTM network.

At each time step $k$, an LSTM cell is composed of several critical components that regulate the flow of information. The input gate $U_k$ is responsible for controlling the amount of new information to be incorporated into the cell memory. This gate ensures that only the relevant input data influences the cell state. The forget gate $f_k$ determines the quantity of existing information in the cell memory that should be discarded, allowing the model to forget irrelevant data and retain essential information. The memory cell $C_k$ functions as the storage unit for the cell state, which is dynamically updated based on the combined influences of the input and forget gates. Finally, the output gate $O_k$ regulates the amount of information to be output from the cell, thereby controlling the influence of the current cell state on the subsequent time steps. These gates operate in a coordinated manner to maintain and manipulate the flow of information within the LSTM cell, enabling the network to capture long-term dependencies and retain pertinent information over extended sequences. The input and output variables of the LSTM network are defined as follows:

$$U_k = \sigma(W_{\Psi u}\Psi_k + W_{hu}h_{k-1} + b_u)$$
$$f_k = \sigma(W_{\Psi k}\Psi_k + W_{hk}h_{k-1} + b_k)$$
$$c_k = U_k tanh(W_{\Psi c}\Psi_k + W_{hc}h_{k-1} + b_c + f_k c_{k-1})$$
$$o_k = \sigma(W_{\Psi o}\Psi_k + W_{ho}h_{k-1} + b_o)$$
$$h_k = o_k tanh_{Ck} \qquad (10)$$

The LSTM cell at each time step $k$ processes several key variables. Here, $\Psi_k$ represents the input layer state at the current time step $k$, while $h_{k-1}$ denotes the hidden layer state from the previous time step $k-1$. The weight matrices $W_{\Psi u}$ and $W_{hu}$ are associated with the input and hidden states for the input gate, respectively, with $b_u$ serving as the corresponding bias term. Similarly, for the forget gate, the weight matrices $W_{\Psi k}$ and $W_{hk}$ are linked to the input and hidden states, and $b_k$) is the bias term. The memory cell $c_k$ is updated based on the influence of both the input gate $U_k$ and the forget-gate $f_k$, incorporating information from the current input and the previous cell state $c_{k-1}$. The output gate, which manages the flow of information from the cell memory to the output, determines the hidden state $h_k$ as a function of the output gate $o_k$ and the current cell state $C_k$. The sigmoid function $\sigma$ is used to ensure the gate outputs are within the range [0, 1], while the hyperbolic tangent function $tanh$ introduces non-linearity and scales the cell state appropriately.

The weight matrices $W$ and bias terms $b$ are crucial for determining the behavior of the gates and the memory cell. The weight matrices $W_\Psi$ denote the connections between the input state $\Psi_k$ and the gates, while $W_h$ represents the connections between the hidden states $h_{k-1}$ and the gates. The bias terms $b$ are added to each gate to allow the model to fit the data more effectively.

The LSTM network is trained using backpropagation through time (BPTT), which adjusts the weights and biases to minimize the error between the predicted and actual outputs [29]. During training, the gradients of the loss function are computed pertaining to the weights and biases, and these gradients are used to update the parameters using optimization algorithms Adam [30].

The structure of the LSTM model for battery data modeling is illustrated in Figure 3. In this model, the inputs include cycle, time, current, voltage, and capacity, which are processed through two-stage LSTM layers. The model outputs the estimated values through the output estimation layer.

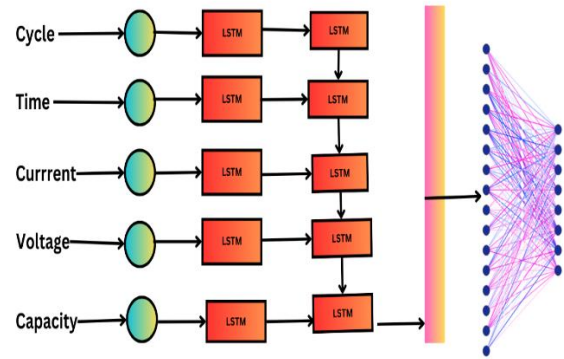

**FIGURE 3.** Architecture of the LSTM model for battery data modelling

### C. CONVOLUTIONAL NEURAL NETWORK

Convolutional neural networks (CNN) are widely used in battery modeling due to their ability to capture spatial and temporal dependencies in structured data [31]. The structured data used in this study consists of voltage, current, capacity, and cycle indices, which are crucial for identifying battery degradation patterns. CNNs effectively extract key features associated with battery health prognosis, enabling accurate prediction of performance and degradation trends over time. The network's capability to process sequential data makes it well-suited for battery state estimation and remaining useful life (RUL) prediction [32], contributing significantly to energy storage system management.

In this model, the input layer is comprised of cycle, time, current, voltage, and capacity as shown in Figure 4. The network's initial layer captures these key attributes, ensuring a structured approach to feature extraction and facilitating efficient learning of temporal dependencies. The input data at the time step $k$ is first passed through a series of 64 one-dimensional convolutional filters within the convolutional layer, which serves as the primary processing unit. These filters systematically slide over the input data to identify essential patterns and features from the cycle data. Mathematically, the convolution operation at the time step $k$ expressed as:

$$Zi,j(k) = \sum_m \sum_n X^{(k)}_{(i-m),(j-n)} W_{m,n} + b \qquad (11)$$

where $X^{(k)}$ represents the input data cycle, time, current, voltage, and capacity at time step k. W denotes the filter weights, b is the bias, and $Z^{(k)}_{i,j}$ is the resulting feature map value at position (i,j).

This activation mechanism enhances the network's ability to learn complex patterns within the sequential data. A Rectified Linear Unit (ReLU) activation function is applied to introduce non-linearity to the model by setting negative values to zero while maintaining positive values, as defined by:
$$f(X(k)) = max(0, X^{(k)}) \quad (12)$$

Subsequently, the pooling layer performs down-sampling operations to retain significant features while reducing the dimensionality of the feature maps. In this model, max pooling is employed to simplify the feature maps and enhance computational efficiency. After the pooling operation, the feature maps are flattened into a one-dimensional vector, preparing them for the fully connected layers of the network.

The fully connected layer processes the extracted features, where each neuron is connected to every neuron in the subsequent layer. This layer integrates the learned representations to make final SOC predictions. The transformation at time step k within the fully connected layer can be expressed as:
$$h(k) = \sigma(Wh^{(k-1)} + b) \quad (13)$$
where $h^{(k)}$ represents the hidden layer state at time step k, W is the weight matrix, and b is the bias term.

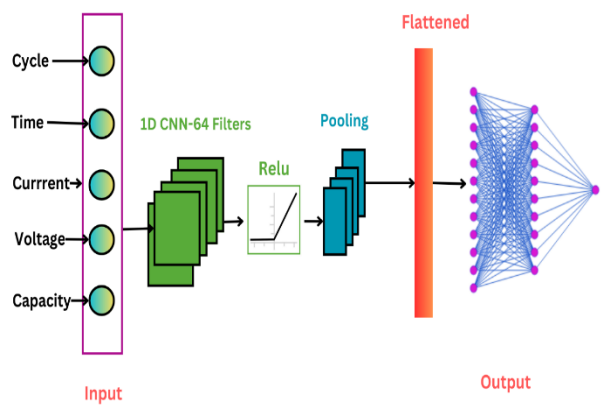

**FIGURE 4.** Architecture of convolutional neural network (CNN)

## D. CONVOLUTIONAL NEURAL NETWORK WITH LONG SHORT-TERM MEMORY

Convolutional neural network (CNN) with Long Short-Term Memory (LSTM) networks' ability to assess important battery metrics such as cycle life, capacity fade, and voltage variations is improved by combining CNN and LSTM architectures. The incorporation of a hybrid model improves the estimation of degradation patterns, allowing for predictive maintenance and increased battery longevity.

The combination of CNN's feature extraction capabilities and LSTM's temporal dependency analysis results in a considerable increase in battery modeling [33]. The CNN component accurately detects complicated patterns in voltage, current, and capacity trends, whilst the LSTM layers track changes across charge-discharge cycles. This hybrid approach greatly improves the accuracy of state-of-health (SOH) and remaining usable life (RUL) estimates.

Furthermore, developments in data augmentation, noise reduction, and feature engineering have improved the model's performance. Extended Kalman Filtering (EKF) and differential capacity analysis techniques improve battery degradation estimation more precisely and reliably [34]. The devised model ensures that real-time monitoring and predictive maintenance procedures may be successfully executed, resulting in increased battery efficiency, lower operational costs, and better energy storage solutions.

CNN consists of numerous layers, each of which performs a different function in feature extraction and input processing. The input layer is the first stage in this design, and it contains raw sequential data. Figure 5 depicts the essential input parameters for this investigation, which are Cycle, Time, Current, Voltage, and Capacity. This model employs a one-dimensional convolutional method with 64 filters, which enables the extraction of local dependencies within sequential data. This process creates feature maps that highlight key patterns in the input. To handle the non-linearity to the model, the Rectified Linear Unit (ReLU) activation function is employed after the convolutional layer [35]. This allows the network to detect complicated patterns and interactions in the data.

After the convolution and activation layers, a pooling layer is used to down sample the feature maps. Pooling decreases the spatial dimensions of retrieved characteristics while retaining the most important information. Common pooling strategies include max pooling, which selects the maximum value from a specific region of the feature map, and average pooling, which calculates the average value within the region. This approach improves computational performance and reduces overfitting [36]. The processed data is fed into Long Short-Term Memory (LSTM) layers, a form of Recurrent Neural Network (RNN) that captures long-term dependencies in sequential data. In this model, the LSTM layers examine the dataset's temporal correlations, efficiently processing sequential information while keeping meaningful patterns across numerous time steps. This integration connects the sequential processing layers to the subsequent fully connected layers.

The LSTM layers have processed the sequence data, the results are converted into a one-dimensional vector. This stage is critical in connecting the sequential processing layers to the next completely connected layers. The final stage of the model is the completely linked layer, which flattens the vector and connects all neurons. This layer combines the information extracted by the convolutional, pooling, and LSTM layers and performs the final computations required for predictive modeling. The output of this layer is used for prediction. The combination of these components results in an efficient and accurate forecast of battery health, allowing for improved battery monitoring and energy storage solutions.

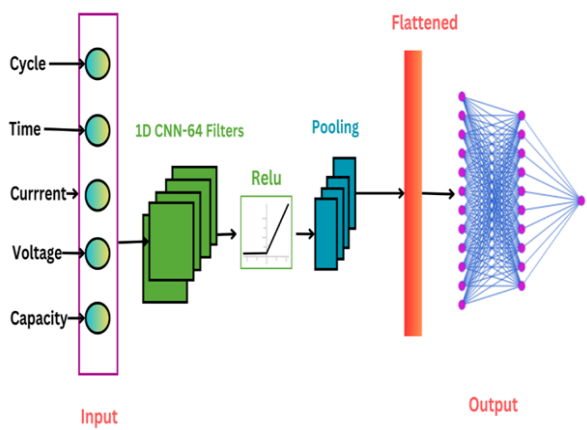

**FIGURE 5.** Architecture of convolutional neural network (CNN) With Long Short-term memory.

## III. EXPERIMENTAL SETUP

A structured battery testing configuration is presented in Figure 6, with the Neware BTS4000 as the core testing system. This eight-channel permits simultaneous testing of multiple lithium-ion batteries under varied conditions. The BTS4000 offers comprehensive investigations of charge/discharge rates, capacity deterioration, and cycle life performance with the maximum rating of 5A/6V, as indicated in TABLE I.

The system involves a host computer, which communicates with the BTS4000 via a TCP/IP protocol, ensuring real-time data acquisition and parameter modifications. Additionally, Ethernet connectivity facilitates remote access, software updates, and system integration for sophisticated analytics. This design provides a stable framework for assessing battery performance across numerous industrial applications, including consumer electronics and electric mobility solutions.

TABLE I
EQUIPMENT SPECIFICATION

| Equipment | Specification |
|---|---|
| Cycling Manufacturer | BTS4000 |
| Cycling tester rating | 5V/6A |
| Test Channel | 0-8 |
| Accuracy | ≥1% |
| Sample frequency | 10Hz |

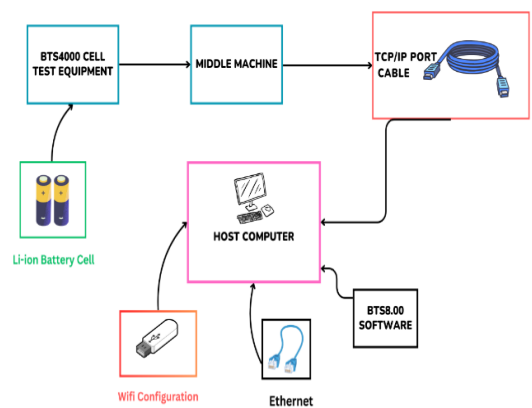

**FIGURE 6.** Experimental test setup for battery test comprised of testing equipment, host computers, and ethernet.

### A. LiNiCoAlO2 BATTERY SPECIFICATION

The battery test was conducted on the A18650 Li-ion battery, manufactured by Hongli. The specifications of this battery, as detailed in TABLE II, include a nominal capacity of 2200mAh, a maximum voltage of 4.2V, a minimum cutoff voltage of 2.5V, and a standard C-rate of 1C (2.2A). These parameters define the battery's performance characteristics and operational limits, which are critical for evaluating its cycle life and degradation behavior under test conditions. The specifications of the LiNiCoAlO$_2$ cell are presented in TABLE II.

TABLE II
BATTERY SPECIFICATION OF A18650 MODEL OF HONGLI LI-ION BATTERY.

| Battery Parameter | Description |
|---|---|
| Model | A18650 |
| Nominal Voltage | 3.7V |
| Nominal Capacity | 2200mAh |
| Maximum Voltage | 4.2V |
| Minimum Voltage | 2.5V |
| Energy | 8.14Wh |
| Cycle Life | 500 |

### B. LiFePo4 BATTERY SPECIFICATION

The ANR266501 lithium-ion battery from A123 Systems operates at a nominal voltage of 3.3V and has a nominal capacity of 2500mAh, signifying its efficiency in charge storage and delivery during a single discharge cycle. To ensure safe charging, the battery is designed with a maximum voltage threshold of 3.6V, preventing overcharging, which could compromise its integrity. Moreover, a minimum voltage cutoff of 2.5V protects against excessive discharge, preserving both longevity and performance. The detailed specifications of the LiFePO₄ cell are presented in TABLE III.

TABLE III
BATTERY SPECIFICATION OF A18650 MODEL OF HONGLI LI-ION BATTERY.

| Battery Parameter | Description |
|---|---|
| Model | ANR266501MB |
| Cell dimension | 26x65 mm |
| Cell weight | 76g |
| Nominal Voltage | 3.3V |
| Nominal Capacity | 2500mAh |
| Maximum Voltage | 3.6V |
| Minimum Voltage | 2.5V |

## IV. METHODOLOGY

### A. BATTERY TEST

Lithium-ion (Li-ion) batteries operate within a specified voltage range, featuring a maximum charge voltage of 4.2V and a discharge cut-off value of 3.0V. While these batteries can be discharged to 2.5V, such action accelerates degradation, diminishes cycle life, and reduces long-term reliability. The charge and discharge rates are regulated by the current rate (C-rate), with protective thresholds established between 2.5V and 4.3V to guarantee safe and reliable testing.

The testing procedure follows a structured sequence, beginning with sample and equipment preparation, followed by initial cell voltage measurement and a load test to determine the appropriate load profile. The battery is then subjected to controlled charge-discharge cycles at varying C-rates, including 0.2C charge with 0.5C discharge, 0.5C charge with 0.9C discharge, 1C charge with 1.3C discharge, and 1.5C charge with 1.6C discharge. These cycles are designed to simulate real-world operating conditions and assess battery performance under different load scenarios.

Once the charge-discharge parameters are established, a cycle life test is conducted for two different *cells*: lithium nickel cobalt aluminum oxide (LiNiCoAlO₂) and lithium iron phosphate (LiFePO₄). Each cycle consists of a 10-minute rest period before charging, followed by charging to the maximum voltage (4.2V for LiNiCoAlO₂ and 3.6V for LiFePO₄), another 10-minute rest period, and subsequent discharge to 3.0V. A final 10-minute rest period is implemented post-discharge to prevent thermal stress and ensure stable operation. For cycle life assessments, a minimum rest period of 20 minutes is introduced between cycles to maintain thermal equilibrium and ensure performance consistency. The structured methodology of this testing framework is shown in Figure 7. This approach ensures a systematic assessment of battery performance, providing valuable insights into degradation mechanisms, cycle life, and overall reliability under varying operational conditions.

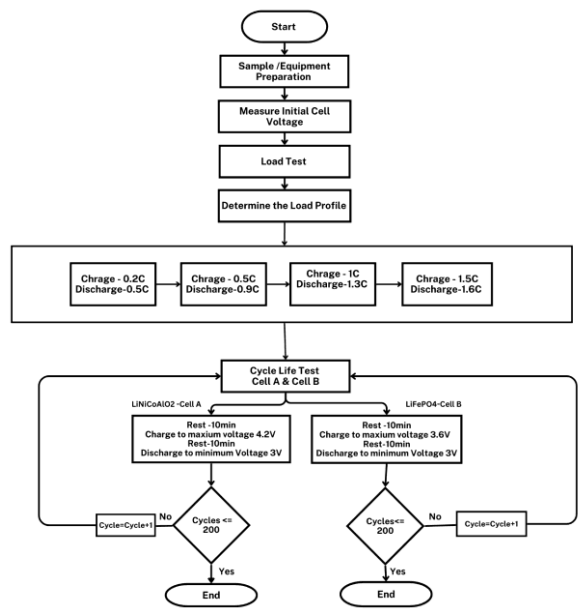

FIGURE 7. Flowchart of the battery cell test design, development, and implementation.

### B. TRAINING STRATEGIES

Datasets for battery cells A, B, C, and D were imported and combined into a single Excel file to start the data processing phase. To ensure consistency in the processing that followed, this stage combined all data sources into a single structured dataset. The time format was standardized to ensure consistency throughout the dataset and remove any inconsistencies that could impact the alignment of battery cycles in terms of time.

Kalman filter is used on the "Current" and "Voltage" columns to improve the data and reduce noise measurement. The filtering procedure successfully reduced noise, resulting in signals that are more accurate reflections of the battery's real performance. After that, we standardized these features on a scale from 0 to 1, which made sure that the model trained steadily and improved its prediction ability.

The dataset is split into training and testing subsets where 70% of the data is allocated for training for sufficient learning and 30% for testing. The data was transformed into an input format suitable for the Long Short-Term Memory (LSTM) network, which requires a timestep of 1 and attributes like current, voltage, and cycle ID due to its sequential nature.

The Keras was used to create the model architecture, which incorporates recurrent and convolutional layers. In the first stage, a 64-filter, one-dimensional convolutional layer (Conv1D) with a 1 kernel size was used to extract spatial features. After that, two long short-term memory (LSTM) layers were incorporated to capture the temporal relationships in the patterns of battery depletion. Each layer has 32 units with ReLU activation. The dropout value was set as 0.2 to improve generalization and avoid overfitting. The architecture was completed with an output layer with a linear activation function.

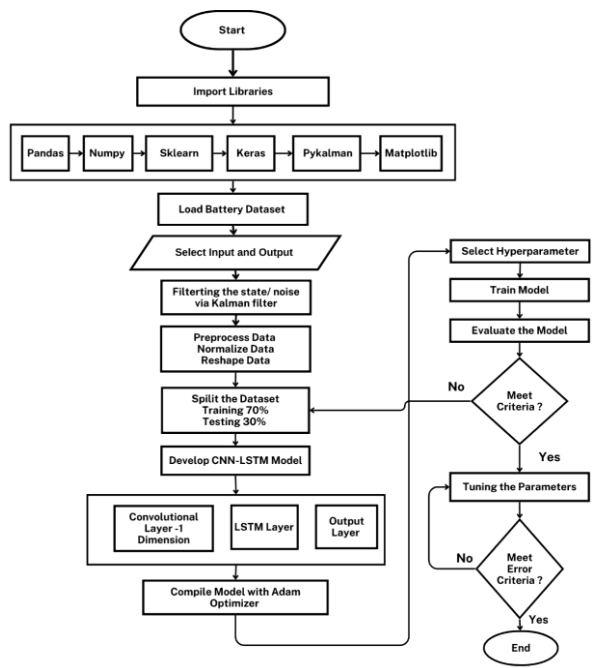

**FIGURE 8.** Flowchart of the developed EKF + CNN-LSTM model.

TABLE IV
TRAINING PARAMETERS OF THE EKF-CNN AND EKF+ CNN-LSTM MODEL.

| Training Parameters | Value |
|---|---|
| Batch Size | 32,64 |
| Epochs | 100,200,300, 400 |
| Learning Rate | 0.001,0.0001,0.00001 |
| Optimizer | Adam |
| Loss function | MSE |

The Adam optimizer was used to compile the model, and the loss function mean squared error (MSE). The model's performance was fine-tuned by hyperparameter tuning as given in TABLE IV, which involved changing important parameters including learning rates (0.001, 0.0001, 0.00001), batch size (32, 64), and the number of training epochs (100, 200, 300, 400) to improve the accuracy. To minimize the error threshold, hyperparameters were fine-tuned until the model matched these criteria. Figure 8 shows the full methodology, which includes data preprocessing and model validation, and provides a structured flow for predicting the health of lithium-ion batteries.

The model performance is evaluated on the error metric performance means squared error (MSE), mean average error performance (MAE), and root-mean-squared error (RMSE). The error metrics performance is stated in the following equations:

$$\text{MSE} = \frac{1}{N}\sum_{k=1}^{N}(C_k - C_k^*)^2 \qquad (14)$$

$$\text{MAE} = \frac{1}{N}\sum_{k=1}^{N}(|C_k - C_k^*|) \qquad (15)$$

$$\text{RMSE} = \sqrt{\frac{1}{N}\sum_{k=1}^{N}(C_k - C_k^*)^2} \qquad (16)$$

Where $C_k$ is the estimated capacity and $C_k^*$ is the experimental capacity at the timestep of $k$.

### C. CELL HEALTH PROGNOSIS MODEL

The differential capacity $\frac{dQ}{dV}$ is a fundamental analytical tool in battery research, which provides insights into electrochemical processes occurring during charge and discharge cycles. The data preparation process begins with importation of measured voltage ($dV$) and capacity ($dQ$) values obtained from cycling tests. To ensure consistency and accuracy, the dataset was arranged in ascending order based on cycle number, with each cycle's data carefully segmented for precise analysis. A capacity vs. voltage plot was then constructed to establish the relationship between capacity and voltage across multiple cycles. This plot served as the basis for calculating the differential capacity $\frac{dQ}{dV}$, which was determined using the differential analysis. To enhance precision, the analysis utilized a dataset of 100 data points, allowing for a more detailed representation of capacity variations with respect to voltage. The cell health prognosis method is detailed in Figure 9.

Due to the inherent noise associated with derivative calculations, the resulting $\frac{dQ}{dV}$ curves exhibited fluctuations, which could obscure significant electrochemical features. To address this issue, a smoothing function was applied to reduce noise and improve the clarity of the differential capacity curve. This refinement enabled the identification of characteristic peaks that correspond to key electrochemical phenomena, such as phase transitions within electrode materials. Identifying these peaks accurately is essential for evaluating the battery's state of charge and understanding the stability of the materials under test.

To ensure reliable peak identification, a peak filtering method was employed with a 30% threshold criterion. This threshold was applied to eliminate minor fluctuations and retain only significant peaks that exhibited a relative prominence of at least 30% of the highest detected peak amplitude. The filtering process involved three key steps: (1) detecting local maxima to identify initial peak candidates, (2) performing baseline correction to ensure peak prominence was accurately assessed, and (3) applying the threshold filtering technique to remove peaks below the 30% prominence level. Following the application of the peak filtering method, key peak parameters, including peak position, peak height, peak area, and peak width, were recorded. These parameters provided a comprehensive assessment of battery performance and internal health, facilitating a deeper understanding of the underlying degradation mechanisms.

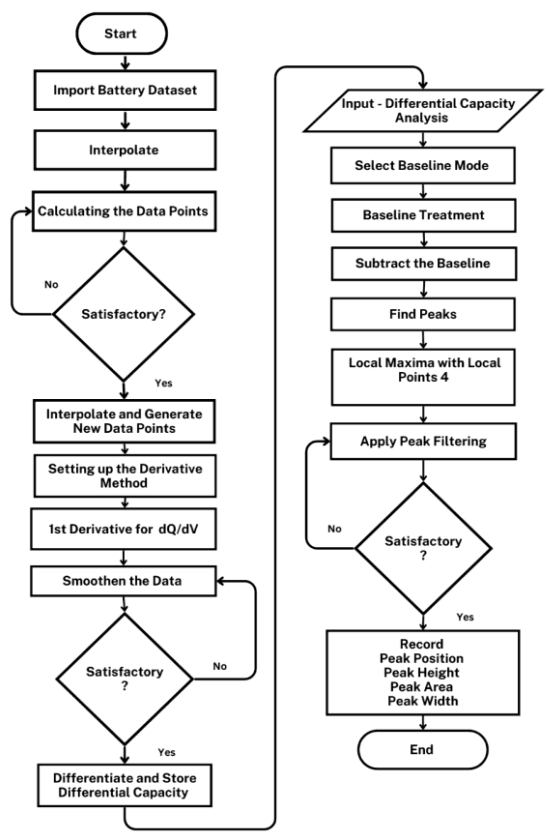

**FIGURE 9.** Flowchart of the battery cell health prognosis utilizing differential capacity analysis and peak identification method.

## IV. RESULT AND DISCUSSION

### A. CELL HEALTH PROGNOSIS MODEL PERFORMANCE

The EKF+CNN-LSTM model's performance during both training and testing phases, evaluated across various C-rates (0.2C, 0.5C, 1C, 1.5C) over 100 epochs, provides a comprehensive understanding of its learning behaviour, generalization ability, and predictive effectiveness. The training and testing performance of the model was analyzed for different C-rates (0.2C, 0.5C, 1C, and 1.5C) over 100 epochs, as illustrated in Figures 10. The training loss exhibits a rapid decline in the initial epoch, indicating effective learning across all cases. However, a distinct pattern emerges for the 0.2C condition, where the initial loss is significantly higher (0.12) compared to the other C-rates, which start around 0.02. This suggests more complex data distribution or increased noise at lower C-rates, requiring additional optimization to minimize the loss. By approximately 30 epochs, the loss stabilizes across all conditions, signifying convergence. Notably, the 0.5C, 1C, and 1.5C curves exhibit a faster convergence rate and reach lower loss values, indicating better adaptability to the training dataset when the data size is smaller.

The testing performance follows a similar trend, confirming the model's ability to generalize effectively. The 0.2C condition again shows a higher initial loss (0.10), followed by gradual stabilization. Despite reaching a near-zero loss, minor fluctuations persist beyond 20 epochs, suggesting potential sensitivity to noise. Conversely, the 0.5C, 1C, and 1.5C curves exhibit smooth and rapid convergence, demonstrating robust generalization capabilities. A close alignment between training and testing losses suggests minimal overfitting, reinforcing the model's effectiveness in capturing the key charge-discharge characteristics across different C-rates.

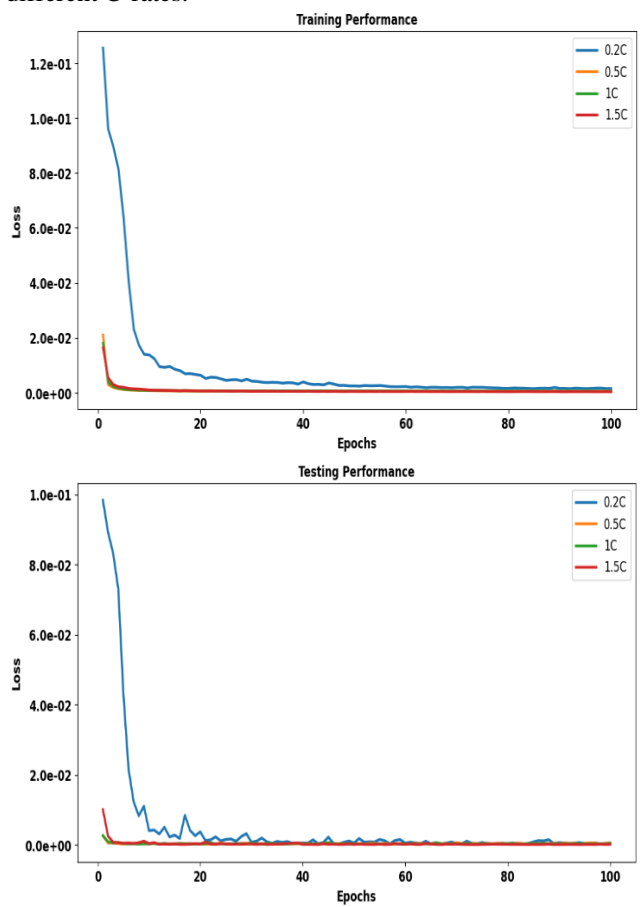

**FIGURE 10.** Training and Testing performance across 100 epochs of EKF+CNN-LSTM model.

The results indicate that 0.2C consistently exhibits the highest loss due to large data accounts and trained initially, while 1C and 1.5C achieve the lowest loss values as the model sufficiently learned data patterns. This trend suggests that higher C-rates facilitate improved learning and improved performance on a smaller dataset. The minimal discrepancy between training and testing losses across all C-rates indicates a well-generalized model, ensuring reliable performance. Overall, the findings confirm that the model demonstrates strong predictive capabilities, with higher C-rates yielding superior convergence and stability.

In both phases, the 0.2C rate presented the model with greater challenges, reflected by higher initial losses due to the noise fluctuations in lower rate charging profiles. However, the model demonstrated significant learning progress, with rapid loss reduction during the initial epochs. In contrast, higher C-rates (0.5C, 1C, 1.5C) consistently exhibited lower initial losses and faster convergence, owing to the more distinct patterns in their data, which facilitated easier learning and prediction.

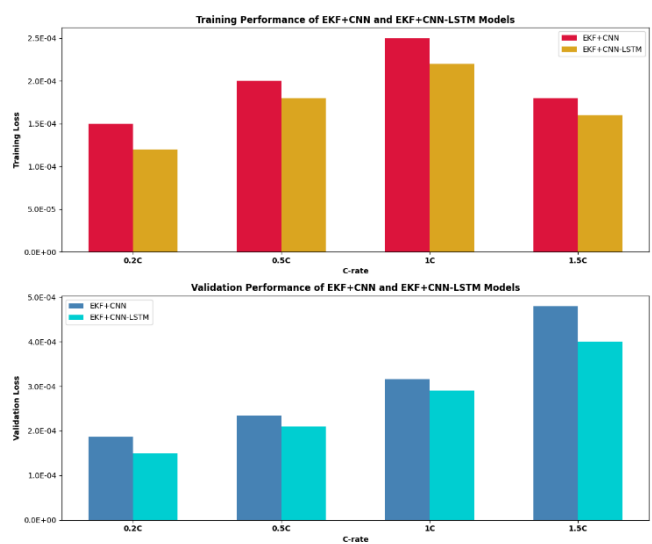

**FIGURE 11.** Training and validation performance comparison between EKF+CNN and EKF+CNN-LSTM model at different C-rates.

The training and validation performances of the Extended Kalman Filter (EKF) integrated with Convolutional Neural Network (CNN) and CNN-Long Short-Term Memory (CNN-LSTM) models were analyzed across various C-rates (0.2C, 0.5C, and 1.5C), as depicted in Figure 11. The training loss, shown in the upper panel, reveals a consistent trend across all C-rates, where the EKF+CNN model exhibits higher training loss compared to EKF+CNN-LSTM. This difference suggests that the inclusion of LSTM layers enhances the model's ability to capture temporal dependencies, thereby improving convergence. At a C-rate of 1C, both models exhibit the highest training loss, with EKF+CNN reaching approximately $2.5 \times 10^{-4}$, while EKF+CNN-LSTM demonstrates slightly lower values. Conversely, at 0.2C, the loss values are significantly reduced, indicating that lower C-rates contribute to a more structured dataset, thereby facilitating improved training stability.

The validation performance, illustrated in the lower panel, follows a similar trend, further confirming the generalization capability of both models. Across all C-rates, the EKF+CNN model consistently exhibits higher validation loss than EKF+CNN-LSTM, reinforcing the latter's superior predictive performance. The highest validation loss is observed at 1.5C, where the EKF+CNN model reaches approximately $4.8 \times 10^{-4}$, whereas EKF+CNN-LSTM maintains a relatively lower loss, indicating improved robustness in handling dataset variations. Furthermore, at 0.2C, both models achieve their lowest validation losses, suggesting that the dataset at this C-rate provides optimal learning conditions. The relatively small gap between training and validation losses across all cases indicates a well-generalized model, with EKF+CNN-LSTM consistently outperforming EKF+CNN due to its ability to preserve long-term dependencies more effectively.

Overall, these findings highlight the advantages of integrating LSTM layers into CNN architectures for battery health prognosis and remaining useful life estimation. While both models demonstrate stable training and validation behaviors, the EKF+CNN-LSTM model consistently outperforms the EKF+CNN model, achieving lower loss values and thereby offering a more accurate and reliable approach for predicting battery degradation under varying operational conditions.

### B. LiNiCoAlO2 CELL PERFORMANCE ANALYSIS DURING CHARGE

The voltage-capacity profiles of CELL A ($LiNiCoAlO_2$), as shown in Figure 12 (left), exhibit a distinct variation across different C-rates (0.2C, 0.5C, 1C, and 1.5C). The observed voltage increase at higher C-rates is a direct consequence of internal resistance and polarization effects, which become more pronounced as current density increases. At 0.2C, the profile follows a gradual and sharp trend. However, at 1.5C, a steeper voltage rise suggests significant polarization, leading to lower charge storage efficiency. This behavior is attributed to diffusion limitations and increased overpotential, which hinder lithium-ion intercalation dynamics. The voltage plateau regions, which are indicative of stable phase transitions, become less prominent as the C-rate increases, suggesting a suppression of reactions at fast charging.

The differential capacity analysis further interprets the impact of C-rate on the electrochemical dynamic of the cell. At 0.2C, five distinct peaks are observed, corresponding to phase transitions associated with $LiNiCoAlO_2$ structural changes. Peak A (3.4V) represents lithium intercalation into the cathode, while Peak B ($\approx$3.6V) is coupled to a secondary phase transition. As the C-rate increases, these peaks become less pronounced and shift towards higher voltages, indicating sluggish phase transformation kinetics and increased reaction overpotentials. The broadening of peaks at higher

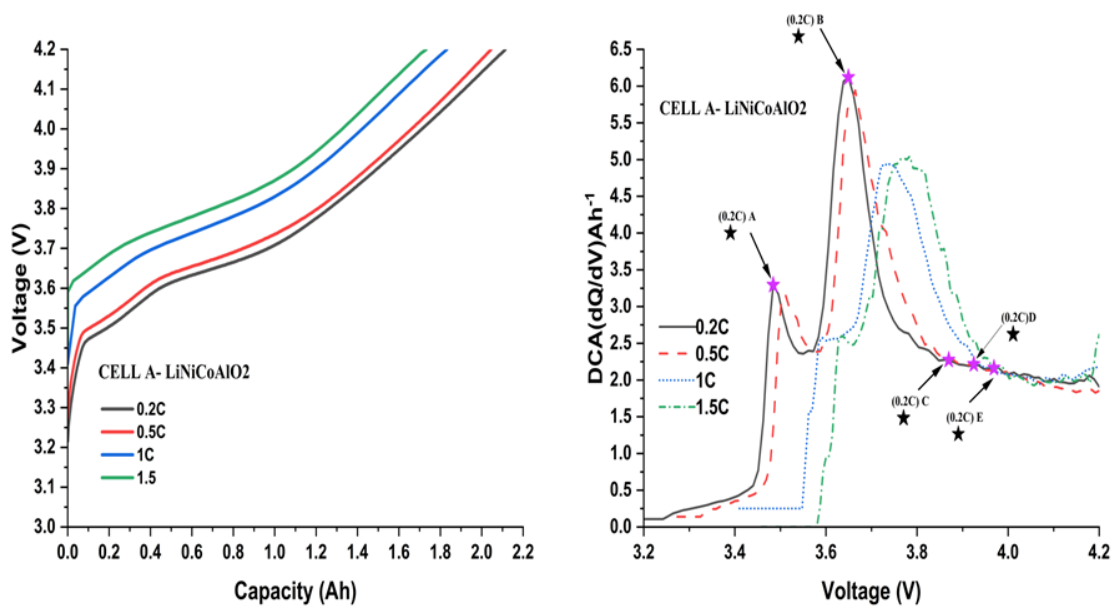

**FIGURE 12. LiNiCoAlO2/graphite cell cycled at different rates in discharge (CHG) profile with Peak labeling.**

C-rates, especially in the intermediate transition regions (C and D), suggests increased ionic resistance and reduced lithium-ion mobility, which accelerates degradation mechanisms such as lithium plating and structural fatigue.

A critical observation is the diminishing intensity of dQ/dV peaks at high C-rates, which signifies reduced electrochemical reversibility. This behavior is particularly concerning long-term battery performance, as suppressed phase transitions lead to capacity fade and irreversible loss of active material. The presence of minor secondary peaks at lower voltages, particularly at 0.2C, suggests a more complete lithiation process, which is partially hindered at higher currents. This aligns with previous studies indicating that rapid cycling accelerates side reactions and structural stress, ultimately compromising the battery's long-term stability.

The findings indicate that while high C-rates enable faster charge-discharge cycles, they come at the cost of increased polarization, suppressed phase transitions, and long-term degradation.

The electrochemical behavior of CELL A under varying C-rates is analyzed through key peak properties: position, height, area, and width. These parameters offer crucial insights into phase transitions, and internal resistance effects during charging, which are essential for understanding performance limitations under high-rate operation.

Figure 13(a) shows that peak positions shift toward higher voltages with increasing C-rate, particularly for peaks C and D, due to increased polarization and overpotentials. Peak A, associated with initial lithiation, exhibits a more gradual shift, indicating lower kinetic constraints. This voltage deviation at high C-rates suggests a long-term capacity fade. As seen in Figure 13(b), peak heights decrease with rising C-rate, particularly for peaks B and C, reflecting reduced electrochemical activity. In contrast, peak A remains relatively stable, implying that early stage lithiation is less affected. The decline in peak height highlights increased charge transfer resistance and diminished reaction efficiency, both of which impact overall battery performance. Figure 13(c) illustrates a substantial reduction in peak area at higher C-rates, particularly for peak B, indicating suppressed charge transfer reactions. A temporary increase at 1C suggests transient diffusion effects before severe polarization dominates. The decreasing peak area correlates with capacity loss, reinforcing the challenges associated with fast charging. Peak width trends, shown in Figure 13(d), reveal significant broadening at high C-rates, particularly for peak B, where width increases from ~150 mV at 0.2C to ~275 mV at 1.5C. This broadening reflects rising polarization and sluggish lithium-ion diffusion. Peak A, however, maintains a relatively stable width, further supporting its lower sensitivity to diffusion constraints.

Overall, these trends underscore the impact of overpotentials, charge transfer resistance, and diffusion limitations at elevated C-rates, leading to structural degradation and capacity loss.

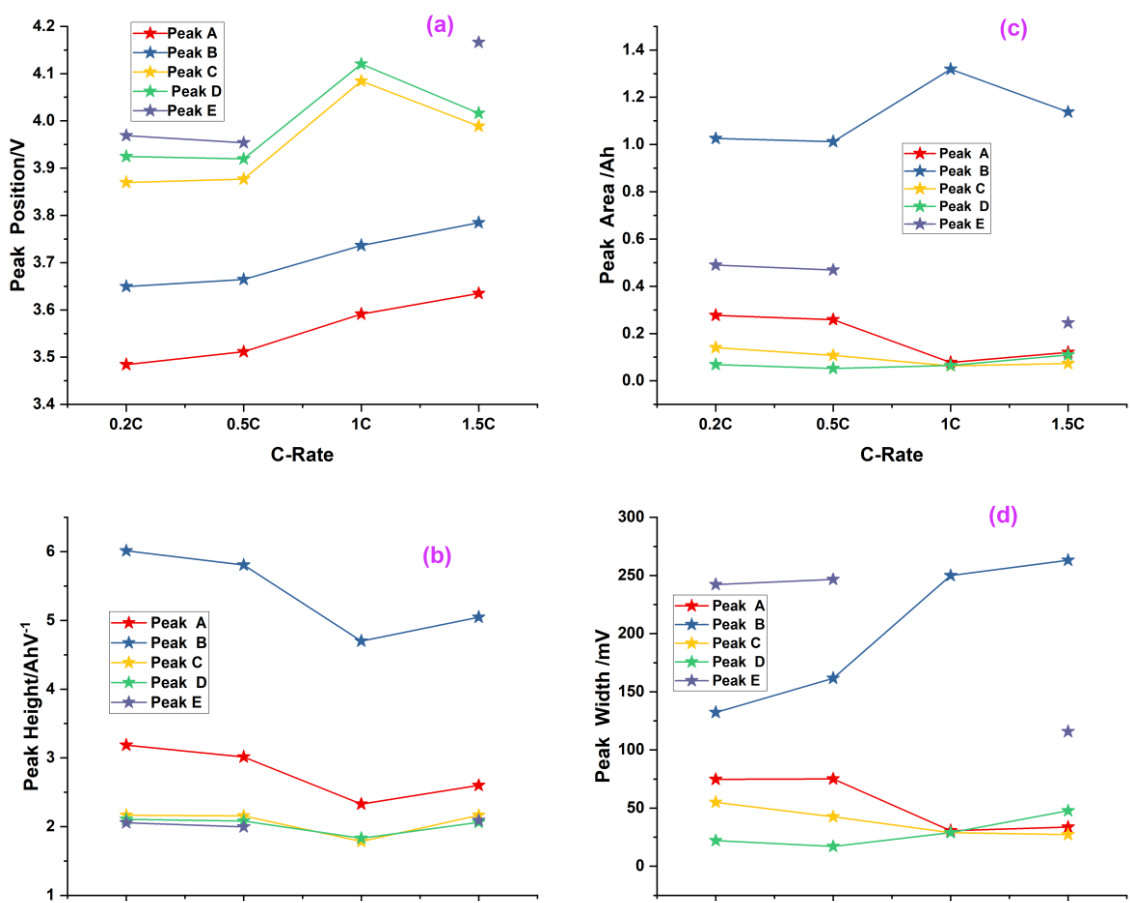

**FIGURE 13.** LiNiCoAlO2/graphite (CELL A) peak properties as a function of C-rate during charge (a) Peak position (b) Peak height and (c) Peak area and (d) Peak width.

## C. LiNiCoAlO2 CELL PERFORMANCE ANALYSIS DURING DISCHARGE

The voltage-capacity profiles of CELL A during discharge in Figure 14 reveal a strong dependence on C-rate, where lower rates (0.5C) result in higher discharge capacities due to minimized polarization and improved lithium-ion diffusion. At higher C-rates (1.6C), a significant capacity reduction is observed, attributed to increased overpotential and internal resistance. The voltage plateau, well-defined at 0.5C, diminishes progressively at 0.9C and 1.3C, becoming nearly indistinct at 1.6C, indicating accelerated lithium deintercalation but with considerable polarization effects.

The differential capacity (dQ/dV) analysis further emphasizes the rate-dependent electrochemical behavior. At 0.5C, sharp and well-resolved peaks correspond to distinct phase transitions during lithium intercalation and deintercalation. However, as the C-rate increases to 0.9C and 1.3C, these peaks shift toward lower voltages and decrease in intensity, suggesting increased polarization and diffusion limitations. At 1.6C, peak broadening and suppression become evident, implying incomplete phase transitions due to transport limitations within the electrode structure.

The results indicate a clear trade-off between power capability and capacity retention in LiNiCoAlO$_2$ cells. While lower C-rates enable higher capacity utilization and well-defined phase transitions, higher C-rates induce polarization effects that hinder lithium-ion kinetics, leading to progressive capacity loss.

The peak properties of the CELL A in Figure 15 exhibit significant variations as a function of C-rate during discharge, providing crucial insights into the electrochemical behavior under different load conditions. The peak position demonstrates a noticeable shift towards lower voltages with increasing C-rate, indicating increased polarization. At 1.6C, the peak position for Peak A drops to approximately 3.35 V, compared to 3.78 V at 0.5C, suggesting a voltage shift of nearly 0.43 V due to higher overpotential and diffusion limitations.

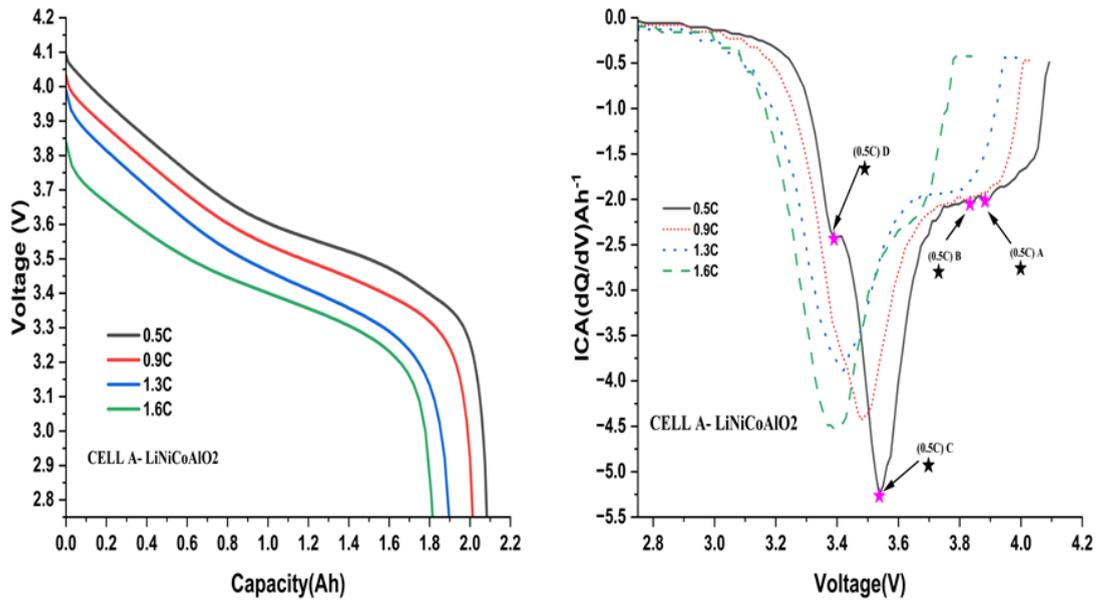

**FIGURE 14.** *LiNiCoAlO2/graphite cell cycled at different rates in discharge (CHG) profile with Peak labelling.*

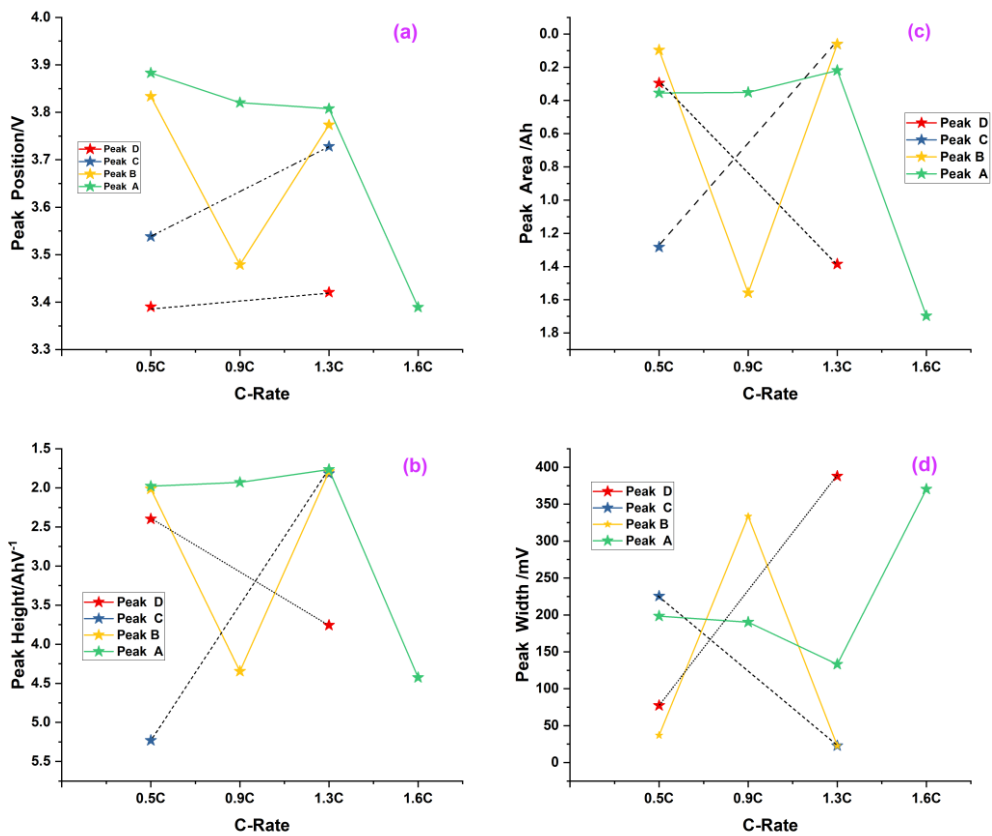

**FIGURE 15.** *LiNiCoAlO2/graphite (CELL A) peak properties as a function of C-rate during discharge (a) Peak position (b) Peak height and (c) Peak area and (d) Peak width.*

The peak height exhibits a strong dependence on C-rate, with Peak A reaching approximately 1.4 A/V at 0.5C but decreasing to nearly 0.6 A/V at 1.6C. This trend signifies increased charge transfer resistance and diffusion constraints at higher C-rates. Similarly, the peak area follows a non-linear trend, with Peak A showing a maximum area of around 1.8 Ah at 0.5C before decreasing sharply to nearly 0.2 Ah at 1.6C, indicating incomplete lithiation processes. The peak width broadens significantly with increasing C-rate, highlighting polarization effects and phase transition broadening. At 0.5C, Peak A exhibits a width of approximately 120 mV, whereas at 1.6C, it expands to nearly 400 mV, reflecting a substantial increase of about 280 mV due to internal resistance.

Overall, lower C-rates facilitate efficient lithium-ion transport and well-defined phase transformations, higher C-rates introduce polarization effects that hinder ion kinetics, reducing performance. These findings highlight the necessity of optimizing discharge rates to balance energy efficiency and power performance in lithium-ion batteries.

### D. LIFEPO4 CELL PERFORMANCE ANALYSIS DURING CHARGE

shortens at higher rates, indicating polarization effects that hinder complete lithium intercalation. The dQ/dV curves show peak broadening and voltage shifts with increasing C-rate. At 0.2C, the peak occurs at ~3.41 V but shifts to ~3.52 V at 1.5C, signifying an increase overpotential. The peak intensity also decreases, indicating slower phase transition kinetics and higher diffusion resistance. The 10% capacity loss and peak voltage shift of ~0.1 V highlight the impact of high discharge rates on electrochemical performance. Increased polarization accelerates degradation, affecting long-term stability. These findings underscore the trade-off between power capability and capacity retention, emphasizing the need for optimized discharge protocols to minimize performance losses.

Figure 17 illustrates the key peak properties for LiFePO$_4$ (Cell A) as a function of C-rate during the charging process, encompassing variations in peak position, peak height, peak area, and peak width. These electrochemical parameters serve as indicators of lithium-ion dynamics, charge transfer characteristics, and polarization effects under varying charge rates. An increase in C-rate results in a notable shift in peak position, with Peak A progressively moving from ~3.34 V at 0.2C to ~3.48 V at 1.5C. This upward trend suggests an

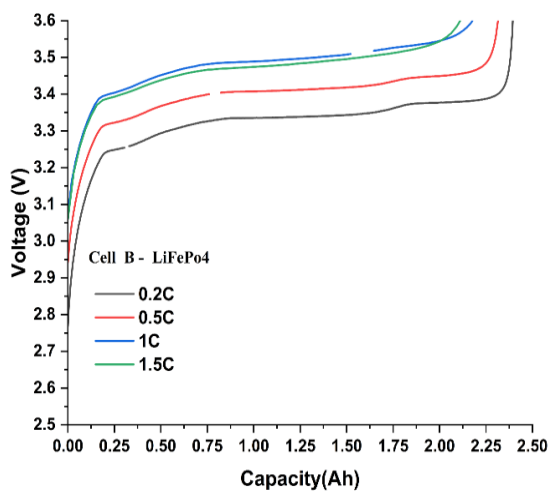
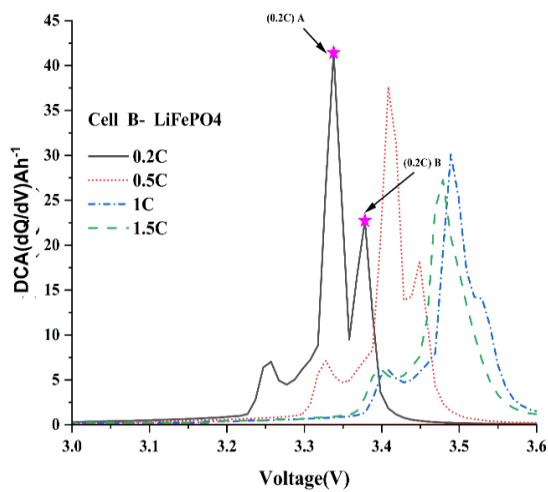

**FIGURE 16.** *LiFePo4 cell cycled at different rates in charge (CHG) profile with Peak labelling.*

The electrochemical behavior of a LiFePO$_4$ (Cell B) was analyzed under charge rates of 0.2C, 0.5C, 1C, and 1.5C as shown in Figure16. The voltage-capacity (V-Q) and differential capacity (dQ/dV) profiles reveal the impact of increasing C-rates on capacity retention, polarization, and phase transition dynamics.

The voltage-capacity curves indicate a decline in capacity with increasing C-rates. At 0.2C, the cell delivers approximately 2.35 Ah, while at 1.5C, the capacity drops by about 10%, reflecting kinetic limitations and increased internal resistance. The voltage plateau around 3.3–3.4 V increasing polarization effect, likely attributable to higher internal resistance and diffusion limitations at elevated charge rates. The observed shift aligns with the expectation that faster charging induces greater overpotential, thereby altering the equilibrium potential of the electrode reaction. Peak height exhibits a declining trend with increasing C-rate, decreasing from ~42 Ah$^{-1}$V$^{-1}$ at 0.2C to ~25 Ah$^{-1}$V$^{-1}$ at 1.5C. This reduction implies a deterioration in charge transfer efficiency, where elevated charge currents impose kinetic constraints on lithium-ion intercalation, leading to increased overpotential and reduced peak intensity. The decreasing peak height is further indicative of an expanded reaction overpotential, which adversely affects the phase transformation kinetics within the electrode material.

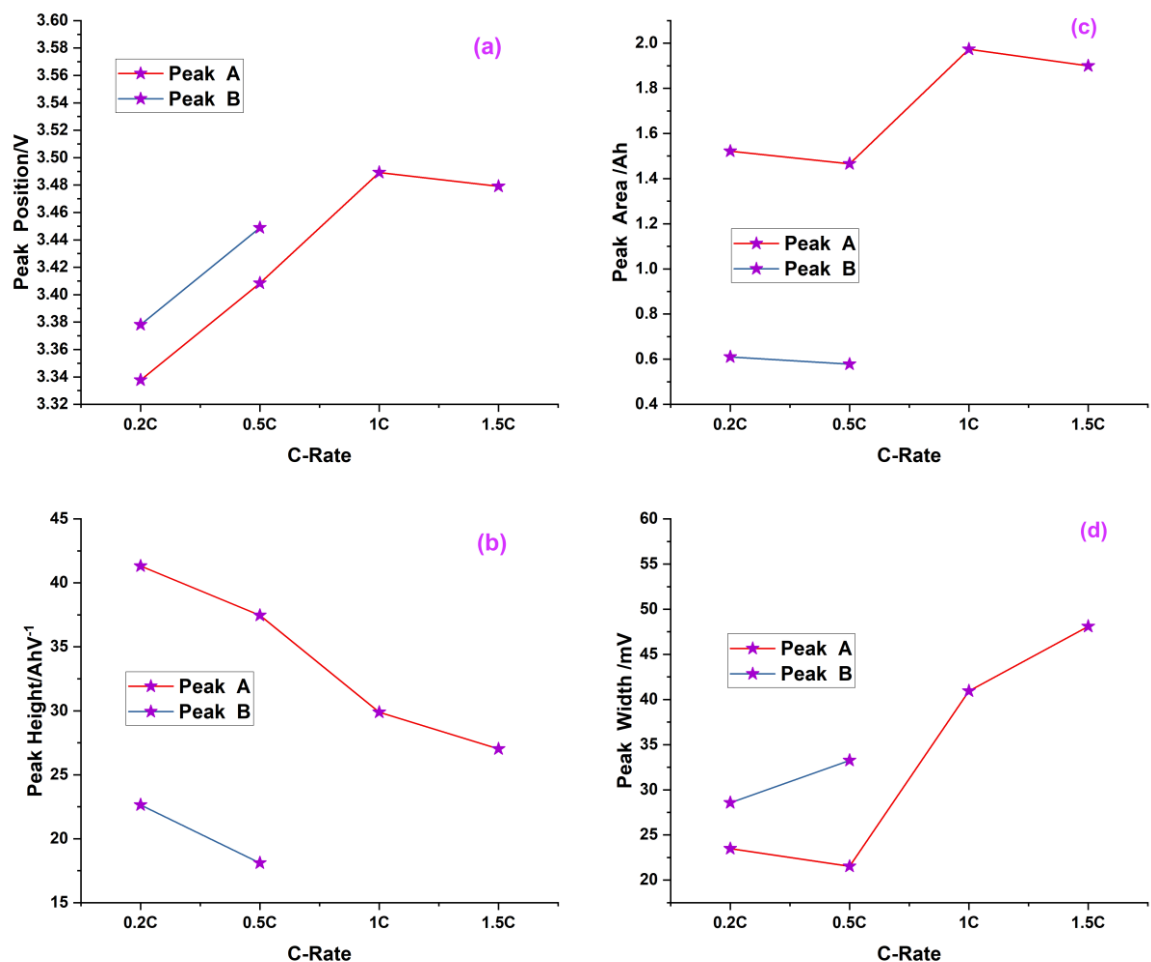

**FIGURE 17.** LiFePo4 (CELL A) peak properties as a function of C-rate during charge (a) Peak position (b) Peak height, (c) Peak area and (d) Peak width.

The peak area, representing the total charge participation in lithium intercalation, remains relatively stable at lower C-rates but increases significantly beyond 1C. This broadening suggests a shift in the charge storage mechanism and increased electrode polarization at higher charge rates. Additionally, peak width experiences substantial broadening, increasing from ~25 mV at 0.2C to ~50 mV at 1.5C. This widening is characteristic of heightened charge-transfer resistance and diffusion impedance, reinforcing the impact of polarization effects under high rate charging conditions. The pronounced broadening at higher C-rates is a consequence of extended reaction overpotential, further emphasizing the trade-off between charge rate and electrochemical stability.

### E. LIFEPO4 CELL PERFORMANCE ANALYSIS DURING DISCHARGE

The discharge characteristics of the LiFePO$_4$ cell in Figure 18 at different C-rates reveal significant variations in voltage response and capacity retention, highlighting the impact of increased current density on cell performance. The voltage-capacity curves demonstrate a progressive decline in the delivered capacity as the discharge rate increases from 0.5C to 1.6C. This behavior is primarily attributed to increased polarization and kinetic limitations, which restrict the full utilization of active material at higher current loads. The characteristic voltage plateau typically observed around 3.2V in LiFePO$_4$ chemistry, becomes less pronounced with increasing C-rate, indicating greater internal resistance and a reduced equilibrium state during lithium intercalation and deintercalation. Additionally, a sharper voltage drop is observed near the end of discharge at higher C-rates, suggesting an accelerated depletion of accessible lithium ions and a greater contribution of ohmic losses.

The differential capacity analysis (DCA), which plots dQ/dV against voltage, provides deeper insights into the electrochemical kinetics and phase transitions occurring within the cell. At 0.5C, two well-defined peaks, labeled (0.5C)-A and (0.5C)-B, are evident, corresponding to the characteristic phase transitions in the LiFePO$_4$ cathode. However, as the discharge rate increases, these peaks exhibit a notable shift toward lower voltages, signifying an increase in polarization effects. Furthermore, the peak intensities progressively diminish with increasing C-rate, reflecting the hindered reaction kinetics and reduced lithium-ion diffusion efficiency at high discharge currents. The broadening of the peaks at higher C-rates further suggests that the phase transition process becomes less distinct due to a loss of thermodynamic equilibrium, leading to a more diffusion-limited intercalation process.

The observed rate-dependent performance degradation has significant implications for the practical application of LiFePO$_4$ in high-power energy storage systems. The increased internal resistance and polarization effects at higher C-rates emphasize the need for improved electrode design strategies to enhance ironic and electronic limitations. This voltage shift suggests that higher discharge rates induce greater overpotential, resulting in an earlier onset of phase transformation. The peak height, representing the intensity of the differential capacity signal dQ/dV, diminishes as the C-rate increases. At 0.5C, the peak height is most pronounced, indicating efficient lithium-ion transport and well-defined two-phase transition behavior. However, with increasing C-rate, the reduction in peak height suggests a more diffusion-limited process, where lithium-ion intercalation occurs at a non-equilibrium state. The suppression of peak intensity at high discharge currents highlights the limitations in reaction kinetics, which can lead to incomplete lithiation and increased electrode polarization. The peak area, which integrates the total charge involved in the phase transition, exhibits a decreasing trend with higher C-rates. A larger peak area at low discharge rates indicates a more extensive phase transformation, where the lithium insertion process occurs more uniformly throughout the electrode. Conversely, as the C-rate increases, the narrowing of the peak area implies that a fraction of the active material becomes underutilized due to transport limitations. This behavior is consistent with the observed reduction in capacity retention at higher discharge currents, reinforcing conductivity. The shifting and broadening of DCA peaks indicate that further investigation into the electrochemical stability of LiFePO$_4$ at high C-rates is necessary to ensure long-term cycle life and efficiency.

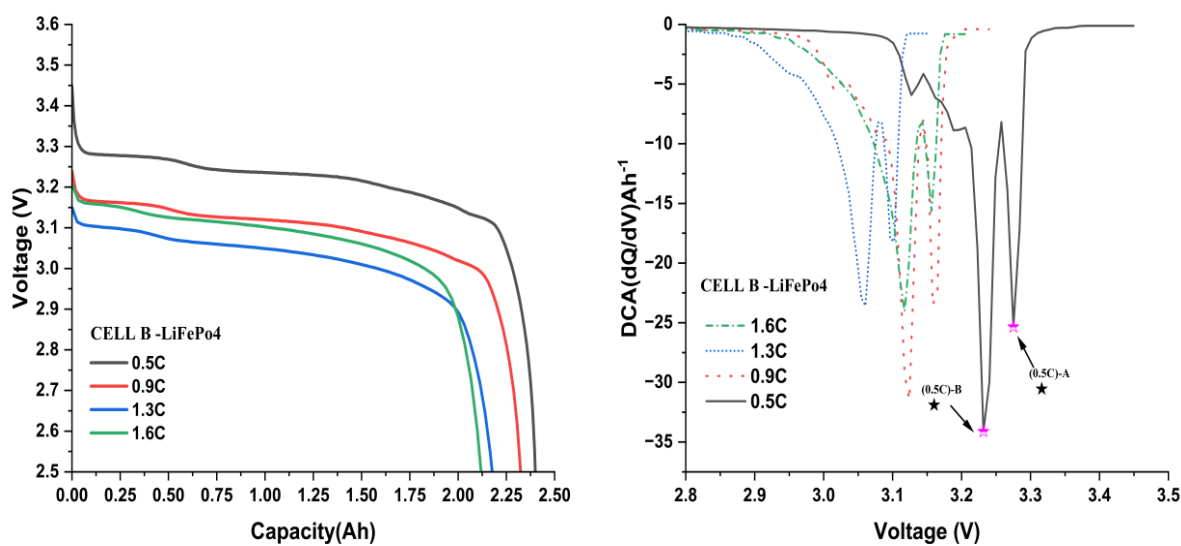

**FIGURE 18.** LiFePo4 cell cycled at different rates in discharge (DCCHG) profile with Peak labelling.

The peak position, which corresponds to the characteristic phase transition voltage during discharge, demonstrates a systematic shift with an increasing C-rate. At lower discharge rates, the peaks are observed at higher voltages, indicating a near-equilibrium lithium insertion process with minimal polarization. However, as the C-rate increases, a distinct shift toward lower voltages is evident, reflecting the growing influence of internal resistance and kinetic the need for enhanced ironic and electronic conductivity in electrode design. The peak width, indicative of the dispersion in phase transition kinetics, shows an inverse relationship with C-rate.

Collectively, these observations underscore the significant

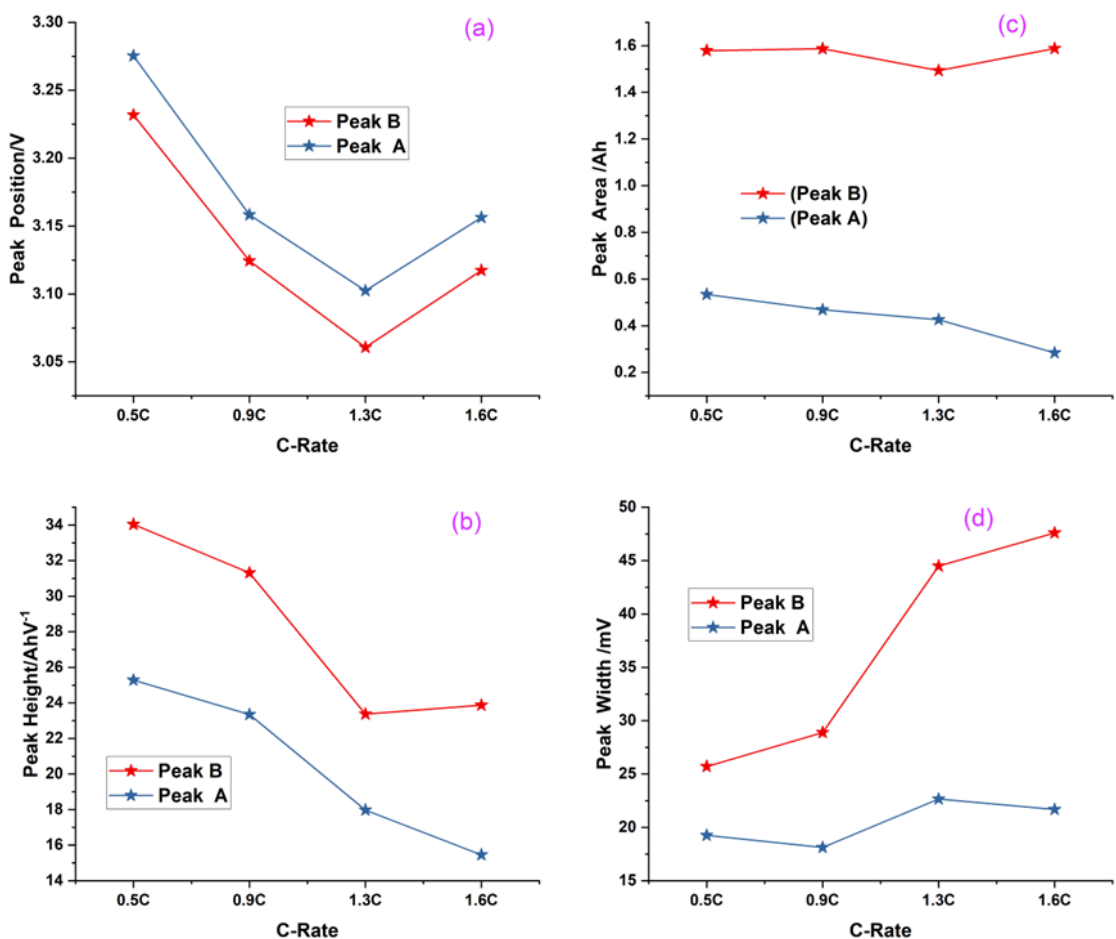

**FIGURE 19.** LiFePo4 (CELL A) peak properties as a function of C-rate during discharge (a) Peak position (b) Peak height, (c) Peak area and (d) Peak width.

At low discharge rates, the well-defined and sharp peaks reflect a clear two-phase equilibrium reaction. However, as the discharge rate increases, peak broadening becomes evident, suggesting that the transition between lithium-rich and lithium-deficient phases occurs over a wider voltage range. This broadening effect is attributed to the loss of thermodynamic equilibrium at high currents, where lithium-ion diffusion constraints lead to a more gradual and less distinct phase transformation.

rate-dependent changes in the electrochemical behavior of LiFePO$_4$. The shifts in peak position, reduction in peak height and area, and broadening of peak width highlight the challenges associated with high-rate discharge conditions, particularly in maintaining optimal lithium-ion transport and minimizing polarization losses. These findings emphasize the necessity of structural modifications, such as particle size reduction, conductive coatings, and electrolyte optimization, to enhance the high-rate capability of LiFePO$_4$-based batteries for practical energy storage applications.

# V. CONCLUSION

The findings of this research underscore the effectiveness of the EKF+CNN-LSTM model in predicting lithium-ion battery health across varying C-rates. A comprehensive evaluation of training and testing performance revealed the model's strong learning capabilities, rapid convergence, and robust generalization. A comparative assessment of EKF+CNN and EKF+CNN-LSTM further highlighted the advantages of integrating LSTM layers. The EKF+CNN-LSTM model consistently exhibited lower training and validation losses across all C-rates, affirming its superior capability in capturing temporal dependencies and maintaining stable learning. The minimal discrepancy between training and testing losses further confirms the model's strong generalization ability, mitigating the risk of overfitting and ensuring reliable performance across diverse operational conditions. Overall, the integration of EKF with CNN-LSTM proved to be a highly effective framework for lithium-ion battery health prognosis. The model demonstrated superior predictive capabilities, particularly at higher C-rates, where it achieved faster convergence and greater stability.

The electrochemical performance analysis of $LiNiCoAlO_2$ and $LiFePO_4$ cells under different C-rates provided critical insights into their charge-discharge dynamics. At higher C-rates, $LiNiCoAlO_2$ exhibited increased polarization effects, resulting in steeper voltage profiles and reduced charge storage efficiency. Differential capacity analysis revealed that phase transition peaks became less pronounced and shifted toward higher voltages, indicating sluggish lithium-ion intercalation kinetics and increased overpotentials. Additionally, peak broadening and reduced intensity at high C-rates suggest accelerated degradation mechanisms, such as lithium plating and structural fatigue, which compromise long-term battery performance. In contrast, $LiFePO_4$ demonstrated greater structural stability and superior cycling performance across varying C-rates. Its voltage profile remained more stable, with minimal polarization effects, suggesting better lithium-ion transport kinetics. Differential capacity analysis showed well-defined phase transition peaks, even at higher C-rates, highlighting its ability to maintain consistent electrochemical performance. Moreover, $LiFePO_4$ exhibited lower degradation rates, reduced overpotentials, and enhanced charge retention, making it a more resilient option for high-power applications. Overall, while $LiNiCoAlO_2$ offers higher energy density, its performance degrades more rapidly at elevated C-rates. $LiFePO_4$, on the other hand, demonstrated superior electrochemical stability and longevity, making it the better choice for applications requiring extended cycle life and high-rate charge-discharge capabilities.

# ACKNOWLEDGEMENT

This project has been partially supported by Gulf University for Science and Technology and the GUST Engineering and Applied Innovation Research (GEAR) Center under project code: ISG – Case 61.

## AUTHOR BIOGRAPHY

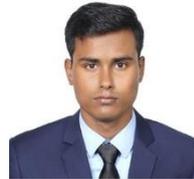

**MD AZIZUL HOQUE** completed his Master of Science degree in Control System Engineering at Universiti Putra Malaysia (UPM) in Serdang, Malaysia in 2025. His pursuit of academic success followed the completion of a Bachelor of Engineering with Honors in Electrical and Electronic Engineering at the Universiti of Putra Malaysia in 2023. Furthermore, Mr. Hoque holds a Diploma in Electrical and Electronic Engineering from Infrastructure University Kuala Lumpur, which he earned in 2018. His academic trajectory demonstrates a strong devotion to higher education, as well as a gradual collection of knowledge and expertise in electrical and electronic engineering. His research focuses on energy storage systems, electric vehicles, charging systems, lithium-ion battery lifetime prediction and health prognosis, artificial intelligence, cyber security, machine learning, and algorithm development to address real-world challenges.

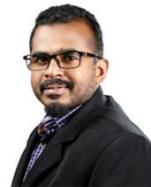

**BABUL SALAM KSM KADER IBRAHIM** earned a PhD in Control Engineering from the University of Sheffield in 2011. He is a highly motivated academic with significant expertise in Electrical & Electronic Engineering, specializing in robotics and control systems, with 15 years of teaching and research experience. Currently, he serves as an Associate Professor in the Electrical and Computer Engineering Department at the College of Engineering and Architecture, Gulf University for Science & Technology (GUST), Kuwait. Prior to this role, he held academic positions in the Faculty of Engineering, Environment and Computing at Coventry University, UK, and in the Faculty of Electrical & Electronic Engineering at University Tun Hussein Onn Malaysia, Johor, Malaysia.

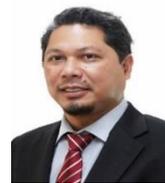

**DR MOHD KHAIR HASSAN** is an Associate Professor in the Department of Electrical and Electronic Engineering at Universiti Putra Malaysia (UPM). He earned his B.Eng. (Hons) in Electrical and Electronic Engineering from the University of Portsmouth, UK, specializing in Control Systems, and completed his Ph.D. in Automotive Engineering at UPM in 2010. With expertise in control systems, automotive control, electric vehicles (EVs), artificial intelligence applications, and the circular economy, Dr. Mohd Khair focuses his research on optimizing energy consumption strategies for EVs, advancing x-by-wire technology, and developing predictive models for lithium-ion battery aging. His innovative work integrates circular economy principles, particularly in second-life applications for batteries, addressing sustainability challenges in energy storage systems and EV technologies. Dr. Mohd Khair has authored numerous high-impact journal articles and serves as a Reviewer for leading academic journals and conferences. He also evaluates research grants, leveraging his expertise to support groundbreaking engineering projects.

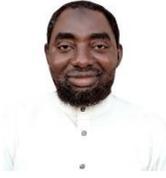 **ABDULKABIR** is a Senior Lecturer at Universiti Teknikal Malaysia Melaka (UTeM) with a distinguished career in engineering and research focused on clean and renewable energy solutions. He holds a PhD in Energy and Power from Cranfield University, United Kingdom, where his research emphasized optimizing biodiesel blends for emission reduction in diesel engines. The research also provided insights into the benefits of hydrogen-fuelled vehicles and electric vehicles. With over two decades of multidisciplinary experience across academia, industry, and government, Dr. Aliyu has contributed extensively to the fields of alternative fuels, emissions analysis, and energy policy. His previous roles include serving as Assistant Chief Scientific Officer at the Energy Commission of Nigeria and as a Research Fellow at Cranfield University. He is actively involved in several funded research projects addressing climate impact, energy storage, and community-based sustainability programs. Dr. Aliyu is a recipient of multiple international awards for research excellence and serves as a peer reviewer for academic journals. He is affiliated with several professional bodies, including the International Association of Engineers, the World Society of Sustainable Energy Technologies, the Nigerian National Committee of the World Energy Council, the Council for the Regulation of Engineering in Nigeria (COREN), and the Nigerian Society of Engineers.

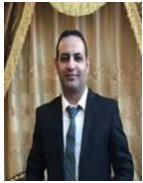 **DR. ABEDALMUHDI ALMOMANY** earned his B.Sc. in Computer Engineering from Yarmouk University in 2005 and went on to complete both his M.Sc. and Ph.D. in Computer Engineering at the University of Alabama in Huntsville in 2015 and 2017, respectively. He began his academic career in 2017 as an Assistant Professor in the Department of Computer Engineering at Yarmouk University. In September 2023, he joined the Gulf University for Science and Technology (GUST) as an Associate Professor in the Department of Electrical and Computer Engineering. Dr. Almomany's research interests lie at the intersection of high-performance computing, FPGA-based reconfigurable systems, embedded systems, parallel processing, and machine learning applications. He has led and contributed to numerous research projects, receiving competitive funding from Yarmouk University, GUST, and the Kuwait Foundation for the Advancement of Sciences (KFAS). These grants have supported his work in developing advanced solutions in high-performance computing and reconfigurable architecture, particularly for real-time and machine learning-driven applications.

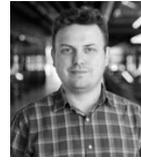 **MUHAMMED SUTCU** received the Ph.D. degree in Industrial and Enterprise Systems Engineering from the University of Illinois Urbana-Champaign, Urbana, IL, USA, in 2014. He is currently working as an Associate Professor with the Department of Engineering Management, Gulf University for Science and Technology, 32093, Hawally, Kuwait. His research interests include decision making with incomplete information and preferences, data-based decision making, information theory, preference elicitation, utility theory, multi-attribute decision making, and simulation modeling, with particular focus on applications in the automobile industry, global warming, and climate change.